\documentclass[11pt]{article}

\usepackage{acl}
\usepackage{microtype}

\usepackage[english,bidi=default]{babel}
\babelfont{rm}{TeXGyreTermesX}
\babelprovide[import]{russian}
\babelfont[russian]{rm}{tempora}

\usepackage{diagbox}
\usepackage{array}
\usepackage{acro}
\usepackage{arydshln}
\usepackage{tabularx}
\usepackage{xcolor}

\usepackage{graphicx}
\usepackage{anyfontsize}
\usepackage{booktabs}
\usepackage{multirow}
\usepackage{colortbl}
\usepackage{rotating}
\usepackage{tikz}    
\usepackage{mdframed}
 
\usepackage{hyperref}

\DeclareAcronym{gpu}{
 short = GPU,
 long = graphics processing unit,
}

\DeclareAcronym{pc}{
 short = PC,
 long = polarity classification,
}

\DeclareAcronym{sc}{
 short = SC,
 long = score classification,
}

\DeclareAcronym{kaznerd}{
	short = KazNERD,
	long = Kazakh Named Entity Recognition Dataset
}

\DeclareAcronym{ner}{
	short = NER,
	long = named entity recognition
}

\DeclareAcronym{nlp}{
	short = NLP,
	long = natural language processing,
}

\DeclareAcronym{asr}{
	short = ASR,
	long = automatic speech recognition
}

\DeclareAcronym{sota}{
	short = SOTA,
	long = state-of-the-art
}

\DeclareAcronym{tts}{
	short = TTS,
	long = text-to-speech
}

\DeclareAcronym{wer}{
	short = WER,
	long = word error rate,
}

\DeclareAcronym{cer}{
	short = CER,
	long = character error rate,
}

\DeclareAcronym{ksc}{
 short = KSC2,
 long = Kazakh Speech Corpus 2
}

\DeclareAcronym{kazakhtts}{
 short = KazakhTTS,
 long = Kazakh Text-to-Speech
}

\DeclareAcronym{kazakhtts2}{
 short = KazakhTTS2,
 long = Kazakh Text-to-Speech 2
}

\DeclareAcronym{kazemotts}{
 short = KazEmoTTS,
 long = Kazakh Emotional Text-to-Speech Synthesis
}

\DeclareAcronym{fleurs}{
 short = FLEURS,
 long = Few-shot Learning Evaluation of Universal Representations of Speech
}

\DeclareAcronym{nes}{
 short = NEs,
 long = named entities
}

\DeclareAcronym{svm}{
 short = SVM,
 long = support vector machine
}

\DeclareAcronym{lr}{
 short = LR,
 long = logistic regression
}

\DeclareAcronym{mnb}{
 short = MNB,
 long = multinomial naïve Bayes
}

\DeclareAcronym{bow}{
 short = BoW,
 long = bag-of-words
}

\DeclareAcronym{tf_idf}{
 short = TF--IDF,
 long = Term Frequency--Inverse Document Frequency
}

\DeclareAcronym{llm}{
 short = LLM,
 long = large language model
}

\title{100,000+ Movie Reviews from Kazakhstan: \\ Russian, Kazakh, and Code-Switched Texts}

\author{Rustem Yeshpanov \\
  Independent Researcher / Astana, Kazakhstan \\
  \texttt{yeshpanov.rustem@gmail.com} \\}

\begin{document}
\maketitle
\begin{abstract}
We present a new publicly available corpus of 100,502 movie reviews from Kazakhstan collected from \texttt{kino.kz}, spanning 2001--2025 and covering 4,943 unique titles. The dataset is multilingual, consisting mainly of Russian reviews alongside Kazakh and code-switched texts. Reviews are manually annotated for language and sentiment polarity, and 11,309 reviews additionally contain explicit user-provided ratings. We define two sentiment tasks—three-way \acl{pc} and five-class \acl{sc}—and benchmark classical \acs{bow}/\acs{tf_idf} baselines against multilingual transformer models (mBERT, XLM-RoBERTa, RemBERT). Experimental results show that transformer models consistently outperform classical baselines on \acl{pc}, while \acl{sc} remains challenging under leakage-controlled evaluation due to severe class imbalance and subtle distinctions between adjacent rating levels.
\end{abstract}

\section{Introduction}

Movie reviews are widely used in sentiment analysis because they contain naturally occurring, explicitly evaluative language and typically provide more context than short social media posts. However, publicly available datasets of movie reviews in Kazakh remain scarce, limiting reproducible research on sentiment modelling in this under-resourced language. In addition, Kazakhstan provides a practically important multilingual setting in which user-generated reviews are predominantly written in Russian, while Kazakh reviews and code-switching also occur.

We introduce a new publicly available corpus of 100,502 movie reviews collected from \texttt{kino.kz}, spanning 25 years (2001--2025) and covering 4,943 unique titles. The dataset includes Russian, Kazakh, and code-switched texts, and is manually annotated for review language and sentiment polarity. A subset of 11,309 reviews additionally contains explicit user-provided ratings, enabling fine-grained score prediction.

We define two supervised sentiment classification tasks: \acl{pc} with three labels (negative/neutral/positive) and \acl{sc} based on user ratings. We report benchmark results for classical \acs{bow}/\acs{tf_idf} baselines and multilingual transformer models, including per-language evaluation to characterise performance under data imbalance and code-switching. The dataset, accompanying documentation, and trained models are released to support future work on multilingual sentiment analysis and culturally grounded user-generated text in Kazakhstan and comparable contexts.

While sentiment classification is not the only possible use of this corpus, it provides a widely understood and reproducible probe task for characterising dataset difficulty and establishing baselines on Kazakhstan-specific review discourse. Beyond aggregate scores, our experiments surface two dataset-specific issues that are easy to miss in cleaner benchmarks: (i) neutral polarity is rare and often expressed implicitly, which leads to systematic confusions, and (ii) fine-grained score prediction is highly susceptible to label leakage because users frequently state ratings verbatim in the text, motivating leakage-controlled evaluation. These baselines therefore serve as reference points for future work on multilingual modelling, code-switching, and robust sentiment inference in real-world review text from Kazakhstan.

\section{Related Work}

Movie reviews are a longstanding benchmark for supervised sentiment analysis, dating back to early polarity-classification work on review corpora \citep{pang-etal-2002-thumbs}. For English, widely used resources such as the IMDb Large Movie Review Dataset \citep{maas-etal-2011-learning,maas2011imdb} and the Stanford Sentiment Treebank \citep{socher-etal-2013-recursive} have enabled extensive comparison of both classical and neural approaches across binary and fine-grained sentiment settings.

For Russian, sentiment datasets exist, but fewer have become standard movie-review benchmarks. A closely related resource is the Kinopoisk movie review corpus \citep{blinov2013research}. Other widely used Russian benchmarks focus on different domains, such as social media (e.g., RuSentiment~\citep{rogers-etal-2018-rusentiment}, and therefore differ from long-form reviews in length, register, and discourse structure.

For Kazakh, publicly available sentiment resources remain comparatively limited. KazSAnDRA \citep{yeshpanov-varol-2024-kazsandra} provides a large-scale Kazakh review dataset (180,064 items) with 1--5 star ratings from four domains (mapping/navigation, e-commerce marketplace, online bookstore, and Android app store). The dataset reflects naturally occurring Kazakh online text, including Kazakh--Russian code-switching and mixed Cyrillic/Latin writing practices, and the accompanying baselines report competitive performance for polarity classification (F$_1$ = 0.81) and substantially lower performance for fine-grained score prediction (F$_1$ = 0.39).

Finally, code-switched sentiment analysis has been studied primarily in short-form social media via shared tasks such as SemEval SentiMix \citep{patwa-etal-2020-semeval}. In contrast, our corpus targets long-form movie-review discourse from Kazakhstan and provides a Kazakhstan-specific multilingual setting; while code-switched reviews form a small subset, they still allow targeted evaluation on naturally occurring Kazakh--Russian mixed-language reviews, within the broader review corpus.

Taken together, prior datasets provide limited coverage for Kazakhstan-specific movie-review sentiment with long temporal span and naturally occurring multilingual (Russian/Kazakh) review text, motivating the dataset release and the use of sentiment baselines as a diagnostic benchmark in this work.

\section{Dataset Development}
\subsection{Source Data}

Movie reviews were collected from \texttt{kino.kz}\footnote{\url{https://kino.kz/}\label{ft:kino}}, a major Kazakh online ticketing and entertainment portal launched in 2000, using BeautifulSoup\footnote{\url{https://www.crummy.com/software/BeautifulSoup}}. The platform allows users to browse showtimes, view trailers, access film information, leave reviews, and purchase e-tickets for films, concerts, theatre performances, sports events, and other cultural activities via both its website and mobile applications (Android and iOS). After removing duplicates, the data collected comprised 100,567 reviews, including review text, review date and author, movie title in Russian/Kazakh and English, screening year, genre, director, duration, age restriction, and production country.

Production-country labels are available for 600 of 4,943 titles (12.1\%). Among titles with known country labels, the most frequent countries (by number of unique titles) are the United States (182), Kazakhstan (110), the United Kingdom (68), Russia (58), and France (48). Kazakhstan is listed as a production country for 18.3\% of titles with known labels. Kazakh-language reviews are more common for these Kazakhstan-produced titles: the median share of Kazakh reviews per title is 0.10, compared to 0.00 for all other titles (mean shares: 0.19 vs 0.009).

\subsection{Review Language Identification and Annotation}

Since the language of the extracted movie reviews was not provided, the author manually identified the language of each review. Unlike~\citet{yeshpanov-varol-2024-kazsandra}, where reviews containing Kazakh-Russian words or grammar were labelled as Kazakh, in this study we aimed to annotate more granularly, labelling reviews as \texttt{kk} for Kazakh, \texttt{ru} for Russian, \texttt{en} for English, \texttt{cs} for instances of code-switching, and \texttt{ot} for all other languages. While \texttt{en} and \texttt{ot} reviews were found, they were extremely rare (65 in total) and were therefore excluded from subsequent analyses. Code-switched reviews include two or more languages within a single text, most commonly Kazakh--Russian, occasionally involving English or other languages.

We distinguish code-switching from loanword usage: the \texttt{cs} label is applied when a review contains a multiword segment from another language (e.g., an inserted phrase or clause), typically including the function words or grammatical marking of that language (i.e., an extended span in the other language). In contrast, isolated conventional borrowings that are integrated into the surrounding language are treated as loanwords and do not, by themselves, warrant \texttt{cs}. Consider the following Kazakh--Russian code-switched review:

\begin{quote}
\foreignlanguage{russian}{\textit{Уақыт аз болмаса, тема фильм. Звуктарды жақсы пайдаланған. 
Сюжет жаксы, но қысқа. Барып көруге стоит.}} \\
\textit{Uaqyt az bolmasa, tema fil'm. Zvuktardy zhaqsy paidalanğan. 
Syuzhet zhaksy, no qysqa. Baryp k\"oruge stoit.} \\
``If you have some time, the film is solid. The sounds are used well.
The plot is good, but short. It is worth going to see it.''
\end{quote}

As Table~\ref{tab:language_sentiment_distribution} indicates, Russian-language reviews constitute the vast majority of the corpus, with smaller subsets of Kazakh and code-switched texts. By whitespace-delimited word count, Russian reviews have a median length of 30 words (95th percentile: 108), Kazakh reviews 24 (95th percentile: 65), and code-switched reviews 33 (95th percentile: 73). The table also shows a strong skew towards positive reviews, a pattern reported for many review platforms~\citep{10.1561/1500000011}; neutral labels are comparatively rare.

\begin{table}[h]
\centering
\fontsize{9}{12}\selectfont
\begin{tabular}{crrrr}
\toprule
\textbf{Language} & \textbf{Negative} & \textbf{Neutral} & \textbf{Positive} & \textbf{Total} \\
\midrule
cs & 201 & 62 & 851 & \textbf{1,114} \\

kk         & 178 & 127 & 2,334 & \textbf{2,639} \\

ru        & 26,936 & 4,140 & 65,673 & \textbf{96,749} \\
\midrule

\textbf{Total}          & \textbf{27,315} & \textbf{4,329} & \textbf{68,858} & \textbf{100,502} \\
\bottomrule
\end{tabular}
\caption{Distribution of movie reviews by language and sentiment polarity}
\label{tab:language_sentiment_distribution}
\end{table}

Moreover, although the platform allows users to rate movies with stars (from one to ten), these ratings are not publicly displayed, complicating the assignment of polarity scores (positive, neutral, negative). Accordingly, the author manually labelled reviews following guidelines specifically devised for this purpose.

In the absence of additional human annotators, we employed gpt-4.1-nano-2025-04-14 as a compensatory measure to support annotation reliability, which was considered a practical solution under the circumstances. The model was instructed as follows:

\begin{mdframed}
\centering
\textit{You are a sentiment classifier for movie reviews (Russian or Kazakh). Return one digit only: 2 -- clearly positive / recommends the movie; 1 -- neutral, mixed, or unclear; 0 -- clearly negative / does not recommend. Always choose one digit; never output anything else.}
\end{mdframed}

\begin{table*}[h]
\centering

\begin{tabular}{c|rrrrrrrrrrr|r}
\toprule
\multicolumn{1}{c|}{\multirow{2}{*}{\textbf{Language}}} & \multicolumn{11}{c|}{\textbf{Score}} & \multicolumn{1}{c}{\multirow{2}{*}{\textbf{Total}}} \\
 & \multicolumn{1}{c}{\textbf{0}} & \multicolumn{1}{c}{\textbf{1}} & \multicolumn{1}{c}{\textbf{2}} & \multicolumn{1}{c}{\textbf{3}} & \multicolumn{1}{c}{\textbf{4}} & \multicolumn{1}{c}{\textbf{5}} & \multicolumn{1}{c}{\textbf{6}} & \multicolumn{1}{c}{\textbf{7}} & \multicolumn{1}{c}{\textbf{8}} & \multicolumn{1}{c}{\textbf{9}} & \multicolumn{1}{c}{\textbf{10}} &  \\
 \midrule
cs & 10 & 2 & 3 & 3 & 2 & 7 & 2 & 8 & 2 & 5 & 106 & 150 \\
kk & 3 & 4 & 1 & 3 & 3 & 5 & 4 & 7 & 10 & 14 & 212 & 266 \\
ru & 357 & 317 & 365 & 568 & 520 & 559 & 1,077 & 699 & 1,244 & 930 & 4,257 & 10,893 \\
\midrule
\multicolumn{1}{c|}{{\textbf{Total}}} & \textbf{370} & \textbf{323} & \textbf{369} & \textbf{574} & \textbf{525} & \textbf{571} & \textbf{1,083} & \textbf{714} & \textbf{1,256} & \textbf{949} & \textbf{4,575} & \multicolumn{1}{r}{{\textbf{11,309}}} \\
\bottomrule
\end{tabular}
\caption{Distribution of user-provided ratings (0--10) by review language \label{tab:language_score_distribution}}
\end{table*}

GPT-generated labels achieved 89.54\% accuracy relative to the single-annotator labels over the full corpus, with substantial agreement (Cohen’s $\kappa = 0.78$), indicating strong consistency beyond chance~\citep{landis1977measurement}. We report these figures to quantify label stability under single-annotator constraints; the released dataset uses the human annotations as the primary labels.

Furthermore, when available, user ratings were extracted from reviews (e.g., 3 out of 10). For reviews where ratings were provided on a 1--5 scale, scores were multiplied by 2 to align with the standard 1--10 scale. In some cases, users explicitly indicated that a movie was so unsatisfactory that it deserved a score of 0, rather than the minimum 1; these instances were accordingly assigned a rating of 0. Consequently, the final rating scale spans from 0 to 10. 

In a small number of cases, the rating format was ambiguous (e.g., a user stating a score of ``3'' without specifying the scale), which could correspond either to 3/10 or to 3/5 (i.e., 6/10 after normalisation). To resolve such cases, we manually inspected the surrounding review content and inferred the most plausible interpretation based on the expressed sentiment. While we applied this procedure consistently and aimed to minimise errors, a limited number of borderline instances may remain, and the extracted scores should therefore be treated as approximate in rare ambiguous cases.

Overall, 11,309 reviews (approximately 11\% of the dataset) contained an explicit user-provided score (e.g., ``10/10'', ``\foreignlanguage{russian}{9 из 10}'' [\textit{devyat' iz desyati}, ``9 out of 10''],  ``\foreignlanguage{russian}{твердая семерка}'' [\textit{tvyordaya semyorka}, ``a solid seven'']). Table~\ref{tab:language_score_distribution} presents the distribution of explicit user-provided scores across review languages.

In addition, during the review inspection, several recurring themes were occasionally noted, such as unmet expectations, whether the movie was a one-time watch, movie sections perceived as unsatisfactory, and cases where the overall impression was negative but the movie was still recommended for niche audiences. While these observations were recorded, they are not the focus of the present analysis.

The language identification and annotation process, carried out single-handedly, spanned 110 days, from August 2025 to January 2026.

\subsection{Collected Data Significance}
\begin{figure*}[t]
  \centering
  \includegraphics[width=\textwidth]{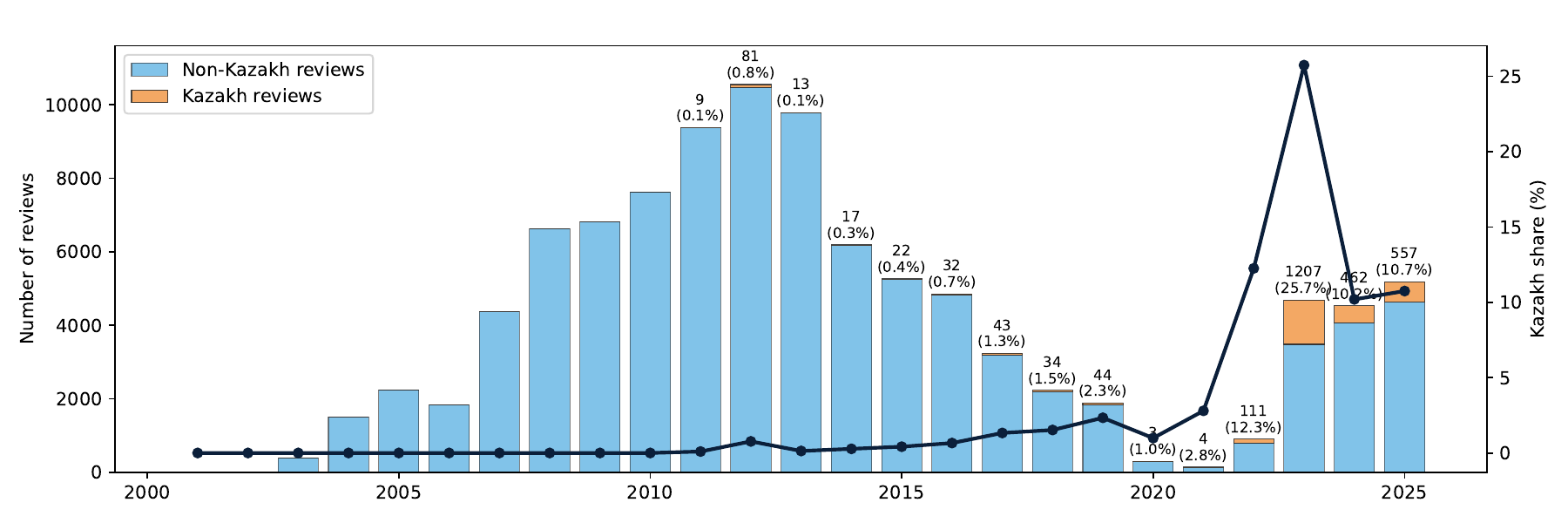}
  \caption{Kazakh-language review share over time}
  \label{fig:kk_share_time}
\end{figure*}

We argue that the collected movie reviews are of substantial value to the \ac{nlp} community for several reasons. First, the dataset spans a period of 25 years, with the earliest reviews dating back to 2001 and the most recent to 2025. Such long-term temporal coverage makes it possible to trace changes in audience preferences and attitudes towards social phenomena and issues (e.g., traditions, domestic violence) over time, ranging from initial denial or avoidance to increased openness and willingness to engage with these topics. Changes in the role and use of the Kazakh language are also clearly observable. In particular, during manual language annotation, we found that although the earliest Kazakh-language review is associated with a film released in 2002, review creation timestamps indicate that the first Kazakh review in our data was authored in 2011, approximately a decade after the launch of the platform (Figure~\ref{fig:kk_share_time}).

This likely reflects the initial predominance of Russian-language reviews and the gradual adoption of Kazakh for user-generated content on the platform. Earlier reviews frequently contain criticism of the quality of Kazakh dubbing and translations, or even explicit requests for permission to express opinions in Kazakh (e.g., \foreignlanguage{russian}{\textit{можно я на казахском}} ``May I speak in Kazakh?''), whereas later reviews increasingly express positive attitudes towards Kazakh-language film production and show greater confidence in using Kazakh to articulate opinions. Notably, the five films with the highest numbers of reviews were all produced in Kazakhstan.

Second, the dataset comprises reviews of 4,943 unique movie titles authored by 31,453 publicly visible reviewer identifiers, reflecting a large and diverse pool of contributors. While many identifiers correspond to self-selected usernames, 6,273 reviews (approximately 19\%) are associated with a generic, platform-assigned label (e.g., ``Kino.kz user'', Russian: ``\foreignlanguage{russian}{Пользователь kino.kz}''), indicating anonymous or non-registered reviewers. Although such entries cannot be distinguished at the individual level, they constitute a substantial portion of the dataset and further contribute to its overall diversity. For release, reviewer identifiers are anonymised by replacing each unique user string with a stable pseudonymous identifier, preserving within-user consistency while removing direct identifiers; reviews associated with the platform-generic label remain indistinguishable, consistent with the source platform.

Third, although the dataset is dominated by Russian-language reviews, the variety of Russian observed is of particular relevance. Specifically, the reviews frequently employ features of Kazakhstani Russian, a regional variety shaped by sustained contact with Kazakh and by local sociocultural context. This includes references to culturally specific events, institutions, and named entities, as well as lexical items and expressions uncommon or opaque to speakers of Russian outside Kazakhstan. Examples include \foreignlanguage{russian}{\textit{ажека, агашка, бастык, токалка, болашаковцы, шапалак, уят, Наурыз, Бауржан Шоу, Sulpak, Керуен, Kcell, Otau Cinema}}, referring to kinship terms, social roles, cultural concepts, holidays, media productions, and local organisations specific to the Kazakhstani context, as well as regionally marked constructions such as \foreignlanguage{russian}{\textit{чёп-чёрный}} (``pitch black''), which illustrates calquing of Kazakh reduplicative intensification patterns into Russian; \foreignlanguage{russian}{\textit{не уятьте}} (``do not shame [someone]''), an example of contact-induced verb formation combining a Kazakh lexical root with Russian negation and imperative morphology; and \foreignlanguage{russian}{\textit{еркеки}} (``men''), formed using a Kazakh lexical root combined with a Russian plural inflection. Such phenomena make the data collected particularly valuable for studying regional language variation, code-switching, and culturally grounded named entity usage in real-world user-generated text.

\subsection{Sentiment Classification Tasks}

Following the design of prior work on Kazakh sentiment analysis, particularly KazSAnDRA, we formulate two primary sentiment classification tasks for our dataset. First, we define a \acf{pc} task, in which reviews are categorised into three broad sentiment categories: positive, neutral, and negative. Second, we consider a \acf{sc} task based on explicit user-provided ratings extracted from reviews.

During dataset construction, user ratings were normalised to a unified 0--10 scale. Accordingly, the initial formulation of the \ac{sc} task involved predicting 11 discrete score labels (0--10). However, as shown in Table~\ref{tab:language_score_distribution}, the distribution of scores is highly imbalanced, with a substantial concentration of reviews assigned the maximum score and relatively few instances in lower score categories. Preliminary experiments with the 11-class setting resulted in unstable training and near-random macro-averaged F$_1$ scores, indicating that the fine-grained formulation is severely affected by data sparsity and long-tailed label distribution.

To obtain more reliable and statistically meaningful results, we therefore adopt a collapsed 5-class \ac{sc} setting, where adjacent score ranges are grouped into broader ordinal bins (0--2, 3--4, 5--6, 7--8, 9--10). Furthermore, due to the highly imbalanced language distribution in the scored subset and the very limited number of Kazakh and code-switched reviews with explicit ratings, the \ac{sc} task is restricted to Russian-language reviews.

\subsection{Data Partitioning}

The data for both tasks were divided into training (Train), validation (Valid), and testing (Test) splits in an 80/10/10 ratio. To reduce topical leakage, splitting was performed at the movie level, so that all reviews of a given film appear in exactly one split. Table~\ref{tab:set_pc_distributions} reports the distribution of reviews across splits by sentiment label and language for the \ac{pc} task. Table~\ref{tab:set_score_sc} reports the distribution of Russian reviews across splits by score bin for the \ac{sc} task.

\begin{table}[!ht]
\centering
\small
\setlength\tabcolsep{0.1cm}
\begin{tabularx}{\columnwidth}{c|rr|rr|rr}
\toprule
\multirow{2}{*}{\textbf{Category}} & \multicolumn{2}{c|}{\textbf{Train}} & \multicolumn{2}{c|}{\textbf{Valid}} & \multicolumn{2}{c}{\textbf{Test}} \\
& \multicolumn{1}{c}{\textbf{\#}} & \multicolumn{1}{c|}{\textbf{\%}}
& \multicolumn{1}{c}{\textbf{\#}} & \multicolumn{1}{c|}{\textbf{\%}}
& \multicolumn{1}{c}{\textbf{\#}} & \multicolumn{1}{c}{\textbf{\%}} \\
\midrule
\multicolumn{7}{l}{\textbf{Sentiment}} \\
\midrule
Negative & 21,767 & 27.21 & 2,890 & 26.69 & 2,658 & 27.50 \\
Neutral  & 3,388  & 4.23  & 502   & 4.64  & 439   & 4.54 \\
Positive & 54,854 & 68.56 & 7,437 & 68.68 & 6,567 & 67.95 \\
\midrule
\multicolumn{7}{l}{\textbf{Language}} \\
\midrule
cs & 809   & 1.01  & 196   & 1.81  & 109   & 1.13 \\
kk & 2,178 & 2.72  & 230   & 2.12  & 231   & 2.39 \\
ru & 77,022& 96.27 & 10,403& 96.07 & 9,324 & 96.48 \\
\midrule
\textbf{Total} & \textbf{80,009} & \textbf{100} & \textbf{10,829} & \textbf{100} & \textbf{9,664} & \textbf{100} \\
\bottomrule
\end{tabularx}
\caption{Distribution of polarity labels and languages across the sets for \ac{pc} \label{tab:set_pc_distributions}}
\end{table}

\begin{table}[!ht]
\small

\begin{center}
\begin{tabularx}{\columnwidth}{c|rr|rr|rr}
\toprule
\multirow{2}{*}{\textbf{Score}} & \multicolumn{2}{c|}{\textbf{Train}}& \multicolumn{2}{c|}{\textbf{Valid}}& \multicolumn{2}{c}{\textbf{Test}} \\
&\multicolumn{1}{c}{\textbf{\#}}&\multicolumn{1}{c|}{\textbf{\%}}&\multicolumn{1}{c}{\textbf{\#}}&\multicolumn{1}{c|}{\textbf{\%}}&\multicolumn{1}{c}{\textbf{\#}}&\multicolumn{1}{c}{\textbf{\%}}\\
\midrule
1 & 839 & 9.56 & 83 & 8.09 & 117 & 10.71 \\
2 & 877 & 9.99 & 116 & 11.31 & 95 & 8.70 \\
3 & 1,349 & 15.37 & 135 & 13.16 & 152 & 13.92 \\
4 & 1,560 & 17.78 & 197 & 19.20 & 186 & 17.03 \\
5 & 4,150 & 47.29 & 495 & 48.25 & 542 & 49.63 \\
\midrule
\textbf{Total} & \textbf{8,775} & \textbf{100} & \textbf{1,026} & \textbf{100} & \textbf{1,092} & \textbf{100} \\
\bottomrule
\end{tabularx}
\caption{Scores across the sets for SC \label{tab:set_score_sc}}
\end{center}
\end{table}

\section{Experiment}

\subsection{Sentiment Classification Models}
For the evaluation of sentiment classification tasks, we employed a set of multilingual transformer-based models that support both Kazakh and Russian and are readily available through the Hugging Face Transformers framework~\citep{Wolf2019TransformersSN}.

\textbf{mBERT}\footnote{\url{https://huggingface.co/google-bert/bert-base-multilingual-cased}} is a multilingual BERT model~\citep{bert} pre-trained on Wikipedia in 100+ languages, including Kazakh and Russian, with a shared WordPiece vocabulary (168M parameters).

\textbf{XLM-RoBERTa}\footnote{\url{https://huggingface.co/FacebookAI/xlm-roberta-base}} is a multilingual RoBERTa model~\citep{DBLP:conf/acl/ConneauKGCWGGOZ20,liu2019roberta} pre-trained on CC-100 CommonCrawl data (100 languages; 270M parameters).

\textbf{RemBERT}\footnote{\url{https://huggingface.co/google/rembert}} is a rebalanced multilingual BERT variant~\citep{chung2021rethinking} trained on Wikipedia and web data in 110+ languages, designed to improve performance on underrepresented languages.

Evaluating instruction-tuned generative models and stronger multilingual encoders is a natural next step. Here we focus on widely used pre-trained multilingual transformers to provide a stable, reproducible supervised baseline that is less sensitive to prompting and instruction-tuning choices.

In addition to transformer-based models, we evaluated several classical machine learning baselines, including linear \ac{svm}, \ac{lr}, and \ac{mnb}, using \acf{bow} and \acf{tf_idf} feature representations, which remain strong and widely adopted baselines for text classification tasks~\citep{Salton1988TermWeightingAI,Joachims1999TextCW}.

\subsection{Experimental Setup}

\subsubsection{Transformer-based models}

All three transformer models were fine-tuned separately for the PC and SC tasks using the corresponding training splits, while hyperparameters were selected based on performance on the respective validation sets. The final model configurations yielding the best validation macro-averaged F$_1$ were subsequently evaluated on the held-out test sets.

Fine-tuning was conducted on \texttt{Vast.AI}\footnote{\url{https://vast.ai/}} using a single NVIDIA RTX 3090 GPU (24~GB VRAM). The total computational cost of all fine-tuning runs was approximately \$2. The initial learning rate was set to $2\times10^{-5}$ and the weight decay to 0.01. Training was terminated early when the validation loss showed a consistent increase across epochs, even in cases where the macro-averaged F$_1$ exhibited marginal fluctuations, in order to mitigate potential overfitting and promote stable generalisation.

Across fine-tuning runs, we used a maximum of three epochs with early stopping based on validation loss. For \ac{pc}, mBERT and RemBERT were trained for two epochs, while XLM-RoBERTa was trained for three epochs. For \ac{sc}, mBERT and XLM-RoBERTa were trained for three epochs, and RemBERT for one epoch. We used a batch size of 150 for mBERT and XLM-RoBERTa on both tasks. Due to GPU memory constraints, RemBERT was fine-tuned with smaller batch sizes (20 for \ac{pc} and 16 for \ac{sc}).

\subsubsection{Classical Baselines}

For all classical models, text was represented using either sparse \ac{bow} or \ac{tf_idf} features with a maximum vocabulary size of 50,000 and n-gram ranges determined via validation-based hyperparameter tuning.

For the \ac{pc} task, the best-performing configuration relied on \ac{bow} features with 1--3-gram representations and lowercasing enabled. Linear \ac{svm} was trained with $C=0.01$ and squared hinge loss, \ac{lr} with $C=2.0$, $\ell_2$ regularisation, and the \texttt{saga} solver, and \ac{mnb} with $\alpha=0.01$ and disabled prior fitting. Class imbalance was addressed using class weights for \ac{svm} and \ac{lr}.

\begin{table*}[!t]
\centering
\small
\begin{tabular}{cccccc|ccccc}
\toprule
\multirow{2}{*}{\textbf{Model}} & \multicolumn{5}{c|}{\textbf{Valid}} & \multicolumn{5}{c}{\textbf{Test}} \\
\cmidrule(lr){2-6} \cmidrule(lr){7-11}
 & \textbf{A} & \textbf{P} & \textbf{R} & \textbf{F$_1$} & \textbf{$\kappa$}
 & \textbf{A} & \textbf{P} & \textbf{R} & \textbf{F$_1$} & \textbf{$\kappa$} \\
\midrule
\multicolumn{11}{c}{\textbf{Polarity classification}} \\
\midrule
\multicolumn{1}{l}{\ac{lr}}             & 0.88 & 0.70 & 0.71 & 0.70 & 0.75 & 0.89 & 0.70 & 0.72 & 0.71 & 0.76 \\
\multicolumn{1}{l}{\ac{mnb}}            & 0.87 & 0.68 & 0.70 & 0.69 & 0.73 & 0.88 & 0.68 & 0.72 & 0.70 & 0.74 \\
\multicolumn{1}{l}{\ac{svm}}            & 0.90 & 0.71 & 0.71 & 0.71 & 0.77 & 0.90 & 0.73 & 0.74 & 0.73 & 0.79 \\
\multicolumn{1}{l}{mBERT}               & 0.92 & 0.78 & 0.70 & 0.72 & 0.82 & 0.93 & 0.81 & 0.72 & 0.74 & 0.83 \\
\multicolumn{1}{l}{RemBERT}             & 0.94 & 0.82 & 0.79 & 0.81 & 0.87 & 0.95 & 0.84 & 0.81 & 0.82 & 0.88 \\
\multicolumn{1}{l}{XLM-RoBERTa}         & 0.94 & 0.80 & 0.78 & 0.79 & 0.86 & 0.94 & 0.81 & 0.81 & 0.81 & 0.87 \\
\midrule
\multicolumn{11}{c}{\textbf{Score classification}} \\
\midrule
\multicolumn{1}{l}{\ac{lr}}             & 0.63 & 0.51 & 0.52 & 0.52 & 0.47 & 0.65 & 0.54 & 0.54 & 0.54 & 0.50 \\
\multicolumn{1}{l}{\ac{mnb}}            & 0.67 & 0.55 & 0.54 & 0.54 & 0.52 & 0.68 & 0.55 & 0.54 & 0.54 & 0.52 \\
\multicolumn{1}{l}{\ac{svm}}            & 0.66 & 0.54 & 0.54 & 0.53 & 0.51 & 0.69 & 0.56 & 0.55 & 0.55 & 0.53 \\
\multicolumn{1}{l}{mBERT}               & 0.64 & 0.52 & 0.52 & 0.50 & 0.48 & 0.67 & 0.53 & 0.51 & 0.51 & 0.50 \\
\multicolumn{1}{l}{RemBERT}             & 0.67 & 0.55 & 0.54 & 0.52 & 0.53 & 0.69 & 0.56 & 0.55 & 0.54 & 0.55 \\
\multicolumn{1}{l}{XLM-RoBERTa}         & 0.66 & 0.51 & 0.55 & 0.52 & 0.52 & 0.68 & 0.55 & 0.56 & 0.54 & 0.54 \\
\bottomrule
\end{tabular}
\caption{Performance of classical and transformer-based models on the \ac{pc} and \ac{sc} tasks}
\label{tab:main_results}
\end{table*}

For the \ac{sc} task, different optimal configurations were observed. Linear \ac{svm} performed best with \ac{tf_idf} features and 1--3-gram representations ($C=0.1$), whereas \ac{lr} and \ac{mnb} achieved stronger results with \ac{bow} features using 1--2-gram and 1--3-gram representations, respectively. The optimal \ac{lr} setup used $C=0.1$ with $\ell_2$ regularisation and the \texttt{saga} solver, while \ac{mnb} was configured with $\alpha=0.5$ and enabled prior fitting. As in the \ac{pc} task, class weights were applied to discriminative models to mitigate class imbalance.

All hyperparameters were selected via grid search on the validation set, with macro-averaged F$_1$ used as the primary optimisation criterion due to class imbalance.

\subsubsection{Score Masking}
For the \ac{sc} task, explicit rating expressions often appear in the review text and may cause label leakage. We therefore replaced all explicit score mentions (identified via manual inspection of the scored subset, focusing on direct numeric and conventional rating expressions) with a placeholder token (\texttt{scoretoken}). The placeholder is alphanumeric for \ac{bow}/\ac{tf_idf} compatibility; it was also added to transformer tokenisers (with resized embeddings) to prevent subword splitting.

\subsection{Sequence Length}

To select an appropriate maximum input length for transformer models, we examined the token length distribution of the training, validation, and test splits after tokenisation with the respective model tokenisers (mBERT, XLM-RoBERTa, and RemBERT). The analysis showed that approximately 95--97\% of reviews contain fewer than 256 tokens, depending on the tokeniser used. Only a small fraction of instances (about 2--5\%) exceed this length and are consequently truncated when a maximum sequence length of 256 is applied.

Considering the relatively low proportion of longer reviews and the quadratic computational complexity of self-attention with respect to sequence length, we set the maximum input length to 256 tokens for all transformer-based models. 

\subsection{Performance Metrics}

We evaluate model performance using accuracy (A), precision (P), recall (R), macro-averaged F$_1$-score (F$_1$), and Cohen’s kappa ($\kappa$). Given the imbalanced class distributions in both \ac{pc} and \ac{sc} tasks, we treat macro-averaged F$_1$-score as the primary evaluation metric, as it assigns equal importance to all classes and offers a more balanced assessment than accuracy alone~\citep{jurafsky2009,yang2001study}. In addition, Cohen’s $\kappa$ is reported to measure the agreement between model predictions and gold labels while accounting for chance agreement.

\subsection{Experiment Results}

We report results for all models in Table~\ref{tab:main_results}.  In general, transformer-based encoders still achieve the highest scores on \ac{pc}, but the gap between them and classical methods narrows after masking explicit score mentions in the \ac{sc} task. Since RemBERT performs consistently well across both \ac{pc} and \ac{sc}, we use it as the main reference system for per-class, per-language, and qualitative error analyses below.

\subsubsection{Polarity Classification}

On the \ac{pc} task, the transformer models continue to set the standard.  The test set macro-averaged F$_1$ scores for RemBERT and XLM-RoBERTa are 0.82 and 0.81, respectively.  mBERT lags slightly behind with an F$_1$ of 0.74.  Among the classical baselines, \ac{svm} and \ac{lr} attain F$_1$ scores of 0.73 and 0.71, respectively, while \ac{mnb} reaches 0.70.  These results indicate that, although simple \ac{bow} models remain strong baselines, contextualised encoders offer a consistent advantage for this three-way classification task.

Per-class analysis for RemBERT in Table~\ref{tab:rembert_pc_per_class} shows uniformly high precision and recall for positive and negative reviews (F$_1$ = 0.97 and 0.94), but markedly lower scores for the neutral class (F$_1$ = 0.56).  The difficulty of detecting neutral sentiment is unsurprising given that neutral instances comprise only about 4--5\% of the corpus and often contain ambiguous language.  The per-language breakdown in Table~\ref{tab:rembert_pc_per_language} also confirms that performance correlates with data volume: Russian reviews dominate the dataset and yield the highest F$_1$ (0.81, $\kappa$ = 0.88), whereas Kazakh (0.77, $\kappa$ = 0.68) and code-switched (0.76, $\kappa$ = 0.73) reviews achieve slightly lower but still respectable results.

\begin{table}[!ht]
\centering
\small
\begin{tabular}{cccc}
\toprule
\textbf{Sentiment} & \textbf{P} & \textbf{R} & \textbf{F$_1$} \\
\midrule
Negative & 0.94 & 0.94 & 0.94 \\
Neutral  & 0.60 & 0.52 & 0.56 \\
Positive & 0.97 & 0.98 & 0.97 \\
\bottomrule
\end{tabular}
\caption{Per-class results of RemBERT on the \ac{pc} test set}
\label{tab:rembert_pc_per_class}
\end{table}

\begin{table}[!ht]
\centering
\small
\begin{tabular}{cccccc}
\toprule
\textbf{Language} & \textbf{A} & \textbf{P} & \textbf{R} & \textbf{F$_1$} & \textbf{$\kappa$} \\
\midrule
cs & 0.87 & 0.87 & 0.74 & 0.76 & 0.73 \\
kk & 0.89 & 0.89 & 0.73 & 0.77 & 0.68 \\
ru & 0.94 & 0.82 & 0.80 & 0.81 & 0.88 \\
\bottomrule
\end{tabular}
\caption{Per-language results of RemBERT on the \ac{pc} test set}
\label{tab:rembert_pc_per_language}
\end{table}

Qualitative inspection of model errors suggests that most confusions involve borderline or implicitly neutral statements. For instance, the Russian review ``\foreignlanguage{russian}{\textit{Неожиданный фильм для меня. Совершенно не похож на воспоминания из детства об индийском кино. Куда делись танцы?!!}}'' [\textit{Neozhidannyy fil'm dlya menya. Sovershenno ne pokhozh na vospominaniya iz detstva ob indiyskom kino. Kuda delis' tantsy?!!}; ``The film was unexpected for me. It is completely unlike my childhood memories of Indian cinema. Where did the dances go?!''] was labelled \texttt{Neutral} but predicted as \texttt{Negative}. Here, the author primarily expresses surprise and a mismatch with genre expectations (the absence of song-and-dance elements typical of Indian cinema), which can be interpreted as either mild criticism or a neutral observation; the model appears to treat the rhetorical question as negative sentiment.

A complementary error appears in a Kazakh review that combines obligation framing with a balanced appraisal: ``\foreignlanguage{russian}{\textit{Озимиздин Казахтар тусиргесин колдап баруга тура келеди, бирак отиниш режиссёр сценаристы кишкене карау керек ау, актерлер оте жаксы ойнап шыкты.}}'' [\textit{Ozimizding qazaqtar tüsirgesin qoldap baruğa tura keledi, biraq ötinish, rezhissior scenaristi kishkentai qarau kerek au, aktiorler öte zhaqsy oinap shyqty}; ``We have to go and support films made by our Kazakhs, but the director and screenwriter really should improve a bit; the actors performed very well.'']. Although the gold label is \texttt{Neutral} due to the combination of critique and praise, the model predicts \texttt{Positive}, likely over-weighting explicit positive cues such as \foreignlanguage{russian}{\textit{``өте жақсы''}} (``very good'') while under-weighting the qualifying criticism. These cases illustrate that neutral reviews in this corpus often encode evaluation indirectly via expectations, rhetorical framing, or mixed praise and critique, making them particularly prone to polarity drift in automatic classification.

\subsubsection{Score Classification}

The \ac{sc} task becomes considerably more challenging under the leakage-controlled setting with masked score mentions.  Across all models, macro-averaged F$_1$ values fall into the 0.50--0.55 range, suggesting that predicting 5-level ratings from text alone is a much harder problem than \ac{pc}.  RemBERT and \ac{svm} achieve the highest test-set F$_1$ of 0.54--0.55, followed closely by XLM-RoBERTa and \ac{mnb} (0.54).  mBERT is the weakest among the transformers with an F$_1$ of 0.51.  The similar performance of classical and transformer models indicates that the masking procedure effectively removed many of the lexical cues that deep models previously exploited, forcing them to rely on more subtle sentiment features in the text.

Per-score analysis in Table~\ref{tab:rembert_sc_per_class} highlights the extreme class imbalance.  RemBERT performs well on the highest rating (score 5) with P = 0.87, R = 0.89, and F$_1$ = 0.88, but struggles on intermediate ratings; the F$_1$ for score 2, for example, is just 0.09 due to very few examples.  This skew explains why overall accuracy is relatively high (69\%) while macro-averaged metrics remain modest. 

\begin{table}[!ht]
\centering
\small
\begin{tabular}{cccc}
\toprule
\textbf{Score} & \textbf{P} & \textbf{R} & \textbf{F$_1$} \\
\midrule
1 & 0.72 & 0.65 & 0.68 \\
2 & 0.24 & 0.05 & 0.09 \\
3 & 0.44 & 0.67 & 0.53 \\
4 & 0.53 & 0.51 & 0.52 \\
5 & 0.87 & 0.89 & 0.88 \\
\bottomrule
\end{tabular}
\caption{Per-class results of RemBERT on the \ac{sc} test set}
\label{tab:rembert_sc_per_class}
\end{table}

\section{Discussion}

The experiments provide two main takeaways. First, multilingual transformer encoders offer a consistent advantage for \ac{pc} on this corpus, but the margin over strong linear baselines is modest. The gap is most visible in the minority neutral class, where language is often ambiguous and underrepresented, while positive and negative sentiment are detected reliably by all models. This suggests that much of the polarity signal in movie reviews can be captured by surface lexical cues, yet contextual modelling remains beneficial for borderline cases and for improving robustness under class imbalance.

Second, leakage-controlled \ac{sc} remains substantially more difficult than \ac{pc}. After masking explicit rating mentions with \texttt{scoretoken}, performance converges across model families and drops to modest macro-averaged F$_1$ values, indicating that fine-grained rating inference depends on subtle and often implicit cues that are harder to learn than coarse polarity. The per-class behaviour further shows that the models perform well on the most frequent high-score bin but struggle on intermediate bins, reflecting both severe label imbalance and the inherently ordinal nature of ratings, where adjacent categories may be expressed with very similar language.

These findings highlight several limitations and directions for future work. The score distribution is strongly skewed toward favourable ratings, and even after collapsing to five bins, the mid-range classes remain sparse. More data in the lower and mid ranges, targeted rebalancing, or modelling approaches that explicitly account for ordinality (e.g., ordinal regression or regression-based formulations) may yield more stable improvements. In addition, fully eliminating score leakage is non-trivial because rating information can be expressed indirectly or idiomatically, and manual score extraction occasionally involves ambiguous cases; consequently, some residual noise in the score labels is likely. Finally, while the polarity results on Kazakh and code-switched reviews are encouraging, stronger conclusions about multilingual generalisation will require more balanced language coverage or dedicated evaluation subsets.

\section{Conclusion}

We introduced a new publicly available corpus of 100,502 movie reviews from Kazakhstan collected from \texttt{kino.kz}, spanning 2001--2025 and covering 4,943 titles, with Russian, Kazakh, and code-switched texts. Reviews were manually annotated for language and sentiment polarity, and 11,309 reviews additionally contain explicit user-provided ratings, enabling fine-grained sentiment modelling.

We defined two supervised tasks: three-way \acl{pc} and five-class \acl{sc}. On \acl{pc}, multilingual transformer encoders achieved the best results, with RemBERT performing strongest (macro-averaged F$_1$ = 0.82, $\kappa$ = 0.88 on the test set). For \acl{sc}, we evaluated a leakage-controlled setting by masking explicit score mentions; under this setup, all models achieved modest macro-averaged F$_1$ scores (0.51--0.55), highlighting the difficulty of inferring rating levels from text alone under severe class imbalance.

To support reproducibility, we release the dataset, annotation guidelines, and trained models on the Hugging Face Hub.\footnote{\url{https://huggingface.co/datasets/yeshpanovrustem/100k_movie_reviews_from_kz}}

\bibliography{custom}

\end{document}